\documentclass{llncs}
\usepackage{makeidx}  
\usepackage{graphicx}
\usepackage{wrapfig}
\usepackage{subfig}
\usepackage{amsmath}
\usepackage{multirow}
\usepackage{array}

\begin{document}
\mainmatter              
\title{Direct Estimation of Regional Wall Thicknesses via Residual Recurrent Neural Network}

\titlerunning{RWT estimation via ResRNN}  
\author{Wufeng Xue, Ilanit Ben Nachum, Sachin Pandey,\\James Warrington, Stephanie Leung, and Shuo Li}
\authorrunning{Wufeng Xue et al.} 
\institute{Department of Medical Imaging, Western University, London, ON, Canada\\
\email{xwolfs@hotmail.com, slishuo@gmail.com}
}

\maketitle              

\begin{abstract}
Accurate estimation of regional wall thicknesses (RWT) of left ventricular (LV) myocardium from cardiac MR sequences is of significant importance for identification and diagnosis of cardiac disease. Existing RWT estimation still relies on segmentation of LV myocardium, which requires strong prior information and user interaction. No work has been devoted into direct estimation of RWT from cardiac MR images due to the diverse shapes and structures for various subjects and cardiac diseases, as well as the complex regional deformation of LV myocardium during the systole and diastole phases of the cardiac cycle. In this paper, we present a newly proposed Residual Recurrent Neural Network (ResRNN) that fully leverages the spatial and temporal dynamics of LV myocardium to achieve accurate frame-wise RWT estimation. Our ResRNN comprises two paths: 1) a feed forward convolution neural network (CNN) for effective and robust CNN embedding learning of various cardiac images and preliminary estimation of RWT from each frame itself independently, and 2) a recurrent neural network (RNN) for further improving the estimation by modeling spatial and temporal dynamics of LV myocardium. For the RNN path, we design for cardiac sequences a Circle-RNN to eliminate the effect of null hidden input for the first time-step. Our ResRNN is capable of obtaining accurate estimation of cardiac RWT with Mean Absolute Error of 1.44mm (less than 1-pixel error) when validated on cardiac MR sequences of 145 subjects, evidencing its great potential in clinical cardiac function assessment.
 
\end{abstract}

\section{Introduction}

\begin{figure}[t]
	\centering
	\includegraphics[width=8cm]{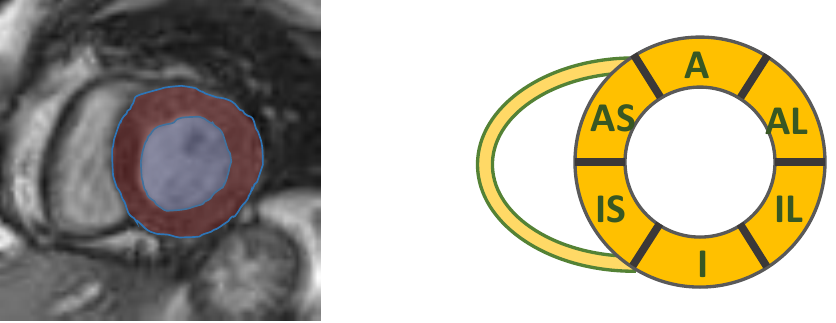}
	\caption{Illustration of RWT for short-axis view cardiac MR image. Left: cardiac image with the contours of myocardium delineated. Right: the 6 segments model for mid-slice cardiac left ventricle. A: anterior; AS: anterospetal; IS: inferoseptal; I: inferior;  IL: inferolateral; AL: anterolateral.}
	\label{fig_RWT}
\end{figure}

Estimation of regional wall thicknesses (RWT) of left ventricle (LV) myocardium is of significant importance for early identification and diagnosis of cardiac disease~\cite{kawel2012normal,puntmann2013left,peng2016review}. Fig.~\ref{fig_RWT} demonstrates the RWT to be estimated for a short-axis view cardiac image. A traditional way for RWT estimation is to segment the LV myocardium from other structures first and then measure the corresponding RWT of each region. However, existing segmentation methods for cardiac images~\cite{ayed2012max,peng2016review,petitjean2011review} require strong prior information and user interaction to obtain reliable results, which may hinder them from efficient clinical application. An alternative way is to circumvent this segmentation step and estimate RWT from cardiac images directly. Direct estimation of cardiac volumes~\cite{wang2015prediction,afshin2012global,afshin2014regional,wang2014direct,zhen2014direct,zhen2015multi,zhen2015direct} have achieved great success in recent years, while direct estimation of RWT has never been explored due to the diversity of cardiac shape and structures for various subjects and various diseases, as well as the complication of regional myocardium deformation through the cardiac cycle.

In this work, we provide a method to estimate the frame-wise RWT from cardiac MR sequences through a newly proposed Residual Recurrent Neural Network (ResRNN). This ResRNN contains two paths: 1) a CNN path for low dimension embedding to robustly represent cardiac images of diverse structures, and preliminary estimation of RWT independently from the embedding of each frame itself, and 2) an RNN path for residual estimation from neighboring frames by leveraging the temporal and spatial dynamic deformation in cardiac sequences simultaneously. In the RNN path, a temporal RNN is applied to the features of temporally neighboring frames for modeling the complex long-range temporal dependencies, and another spatial RNN is applied to the predicted results of spatially neighboring wall thicknesses for modeling the mutual dependencies among these wall thicknesses. For the RNN module, a new Circle-RNN is designed to eliminate the effect of null initial hidden states by characterizing the periodicity of cardiac sequences and the circular spatial layout of cardiac RWT. With image represented by the CNN embeddings, the dynamic deformation of myocardium and the diversity of cardiac shape are well captured by the temporal and spatial RNN, thus leading to accurate estimation of RWT.

\subsection{Related work}

\subsubsection{Segmentation-based and direct methods for cardiac volumes estimation.} Currently, the most related work to RWT estimation is cardiac volumes estimation, which falls into two categories: segmentation-based methods~\cite{ayed2012max,peng2016review,petitjean2011review} and direct methods~\cite{wang2015prediction,afshin2012global,afshin2014regional,wang2014direct,zhen2014direct,zhen2015multi,zhen2015direct}. Segmentation-based methods rely on the premise of cardiac segmentation, which is still a great challenge due to the diverse structure of cardiac image and therefore requires strong prior information and user interaction~\cite{ayed2012max,peng2016review,petitjean2011review}. 

To circumvent these limitations, direct methods without segmentation have grown in popularity in cardiac volumes estimation~\cite{wang2015prediction,afshin2012global,afshin2014regional,wang2014direct,zhen2014direct,zhen2015multi,zhen2015direct}. In these methods, hand-crafted features extracted from cardiac images are fed into regression models such as random forest (RF), adaptive K-clustering RF (AKRF),  Bayesian model, and neural networks, to predict cardiac volumes. The employed features include Bhattacharyya coefficient between image distributions~\cite{afshin2012global,afshin2014regional}, appearance features~\cite{wang2014direct}, multiple low level image features~\cite{zhen2014direct}, as well as features from multiscale convolutional deep belief network (MCDNB)~\cite{zhen2015multi} and supervised descriptor learning (SDL)~\cite{zhen2015direct}.  Although these methods obtained effective performance, two limitations still exist: 1) they followed two separated phases, i.e., \textit{feature extraction}+ \textit{volumes regression}, and no feedback exists between them to make them compatible with each other; 2) neither the temporal dependencies nor the spatial dependencies are taken into account, while the dependencies are important for dynamic modeling of cardiac sequence. In the present work, we provide an elegant solution for direct RWT estimation with an end-to-end architecture that respects both temporal and spatial dependencies.

\subsubsection{Recurrent neural network.} 

Recurrent neural network, especially when the long short-term memory units (LSTM) are deployed, is specialized in long-range temporal dynamic modeling and spatial context modeling. Promising results have been achieved by RNN in a wide spectrum of applications including language modeling~\cite{zhang2015tree}, object recognition~\cite{liang2015semantic}, visual recognition and description~\cite{donahue2015long,li2016online}, and also medical image analysis~\cite{shin2016learning,kong2016recognizing,poudel2016recurrent}. In the work of cardiac image segmentation~\cite{poudel2016recurrent}, an RNN was employed to capture the spatial changes of cardiac structure (represented as low dimensional CNN embeddings) in cardiac sequences. In~\cite{kong2016recognizing}, an RNN with LSTM was employed to model the temporal dependencies in cardiac MR sequences to identify the end-diastole and end-systole frames across a cardiac cycle. In~\cite{shin2016learning}, an RNN was trained to describe the contexts of detected disease in Chest X-Rays. These methods only explored one of the spatial or temporal dependencies while in cardiac sequences, both are important for robust dynamic modeling.

To fully explore the dependencies that exist in cardiac sequences during RWT estimation, two RNN modules are deployed in our work accounting for the temporal and spatial dependencies simultaneously. Besides, we propose a Circle-RNN for periodic cardiac sequences to better serve this aim, avoiding the effect of the null initial hidden input in existing RNN.

\subsection{Contributions}
The contributions of our work include:

\begin{itemize}
	\item An effective end-to-end method that has great potential in clinical cardiac function assessment is proposed for direct cardiac RWT estimation, which has never been investigated previously. 
	
	\item The newly proposed two-path ResRNN endows the network with abilities to robustly represent complex cardiac structure, and effectively model the capricious spatial and temporal dynamic deformation.  

	\item The temporal RNN that accounts for the temporal deformation of the cardiac shape, and the spatial RNN that accounts for the smoothness of the LV myocardium shape, enable ResRNN to estimate collaboratively RWT of all frames and all regions by leveraging the temporal and spatial dependencies in cardiac MR sequences, rather than to estimate independently each cardiac RWT from a single image. 

    \item A Circle-RNN designed for characterizing the periodicity of cardiac sequences and the circular spatial layout of cardiac RWT is proposed to eliminate the effect of the null hidden input for the first time step, to incorporate both the future and the past information in the dynamic modeling, and to treat every frame in the sequence with equally long-term dependencies.	
\end{itemize}

\section{RWT Estimation via ResRNN}

\subsection{Problem formulation}
For a set of cardiac MR sequences $\mathcal{X}=\{X^{s}_{f}\}$, where $s=1\cdots S$ indexes the subject and $f=1\cdots F$ indexes the frame in a cardiac cycle, we aim to estimate the frame-wise values of RWT $\mathcal{Y}=\{y^{s}_{f,l}\}$ for all the frames, where $l=1\cdots 6$ indexes the spatial location of each RWT (see~Fig.\ref{fig_RWT}, from IS to AS in counter-clockwise order). We consider in this work the mid-cavity of LV myocardium in short axis view, which is divided into six segments according to the AHA 17 segments model~\cite{cerqueira2002standardized}. The objective function can be formulated as:
\begin{equation}
\min_{\theta} \frac{1}{2S\times F}\sum_{f}\sum_{s}\|y^{s}_f-\mathbf{Q}(X^s_f|\theta)\|^2_2+\lambda \mathcal{R}(\theta)
\end{equation}
where $\mathbf{Q}$ is the network, $\mathbf{\theta}$ is the parameter vector to be learned, and $\mathcal{R}(\theta)$ regularizes the parameter vector.

\begin{figure}[h]
	\centering
	\includegraphics[width=10cm]{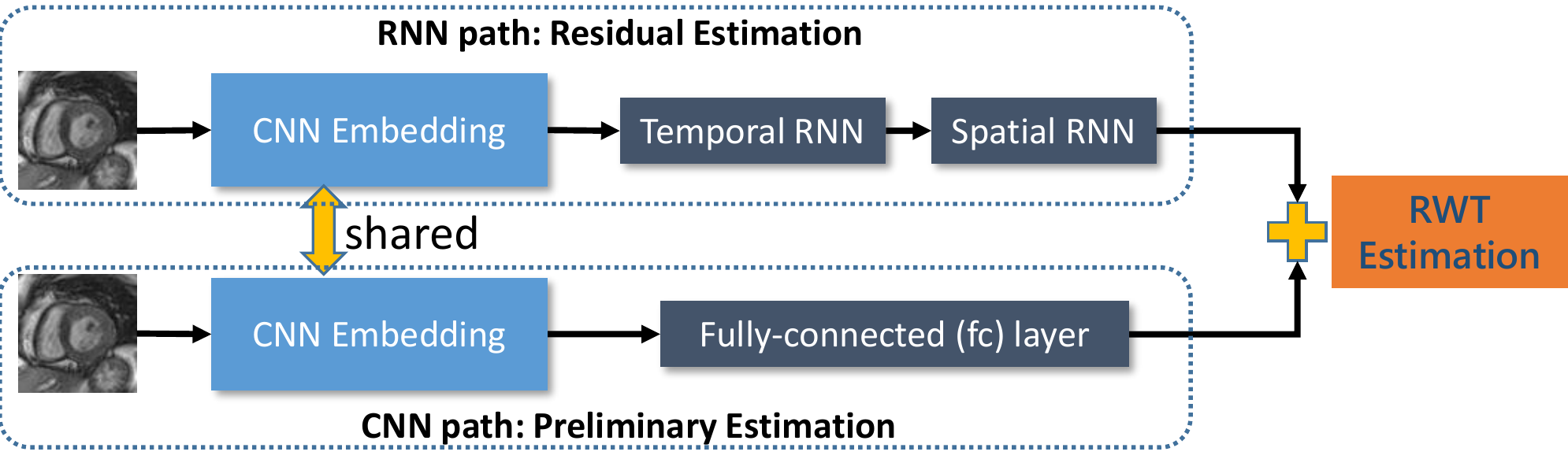}
	\caption{Overview of ResRNN. A CNN path (the lower part) is employed to learn CNN embedding for the cardiac image and further obtain a preliminary estimation of RWT from these features. An RNN path (the upper part) is employed to model the temporal and spatial dependencies based on the CNN embedding of the cardiac sequences to estimate the residual of the preliminary estimation.}
	\label{fig_overview}
\end{figure}

\subsection{Overview of the proposed method}
\noindent To estimate the frame-wise RWT from cardiac MR sequence, we build a new architecture of network: ResRNN. As shown in Fig.~\ref{fig_overview}, two paths are included in ResRNN: with the CNN path $\mathbf{Q_{CNN}}$, each frame in the sequence is independently processed by the CNN network, forming a low dimensional embedding of the cardiac images, from which the RWT is preliminarily estimated with another fully-connected layer; with the RNN path $\mathbf{Q_{RNN}}$, two RNN modules are deployed to model the temporal dependencies between neighboring frames and the spatial dependencies between neighboring RWTs, so as to further correct the residual of the preliminary estimation. The RNN path shares the same CNN embedding with the CNN path. The final RWT estimation is computed as:
\begin{equation}
\mathbf{Q}(X^s_f|\theta)=\mathbf{Q}_{CNN}(X^s_f|\theta)+\mathbf{Q}_{RNN}(X^s_f|\theta)
\end{equation}

\begin{figure}[th]
	\centering
	\includegraphics[width=10cm]{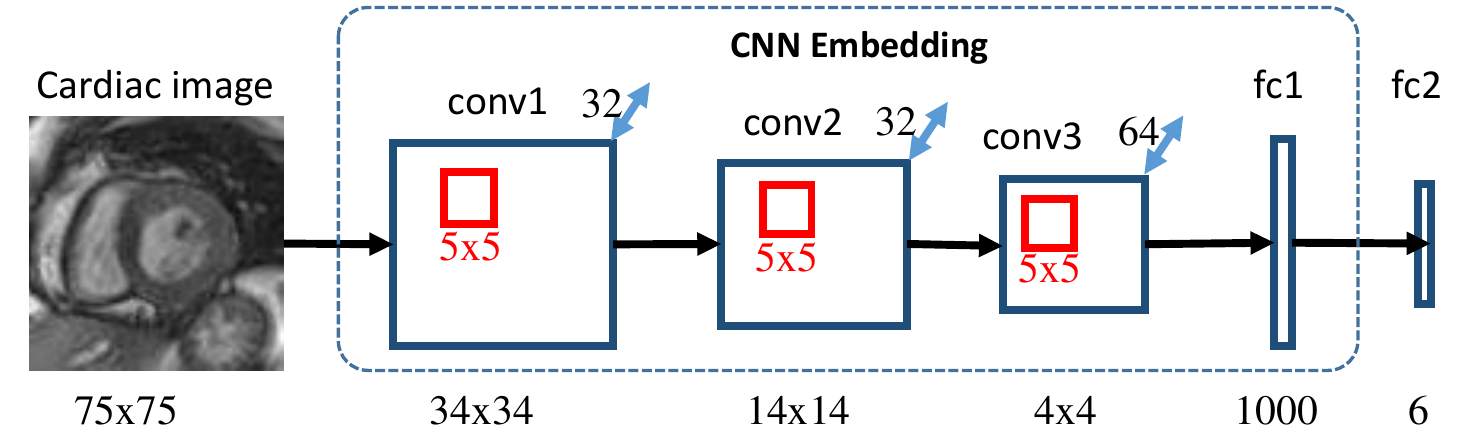}
	\caption{Diagram of the CNN path. Each convolution layer (conv) is followed by a rectified linear unit (ReLU) layer and a max-pooling layer. The CNN embedding part represents the cardiac image as a low dimensional feature vector, which is shared by both the CNN and RNN paths.}
	\label{fig_cnn_branch}
\end{figure}

\subsection{Preliminary estimation with the CNN path}
\noindent The diagram of our CNN path is shown in Fig.~\ref{fig_cnn_branch}. Three convolution (conv$1\sim3$) and one fully-connected (fc1) layers are deployed to obtain the low dimensional CNN embedding $e^s_f$ of cardiac images. The second fully-connected layer (fc2) estimates a preliminary results $y^{s,CNN}_f$ of RWT from the CNN embeddings. 
\begin{equation}
y^{s,CNN}_f=\theta_{fc2:w}e^s_f+\theta_{fc2:b}
\end{equation}
where $\theta_{fc2:w}$ and $\theta_{fc2:b}$ are the weight matrix and bias of the fc2 layer.

As a feed forward neural network, the CNN path bears a notable limitation that it relies on the assumption of independence among samples, which does not hold for cardiac sequence. The dependencies among cardiac sequences have to be modeled to further reduce the residual of the CNN estimation.

\subsection{Residual estimation with the RNN path}
\begin{figure}[th]
	\centering
	\includegraphics[width=12cm]{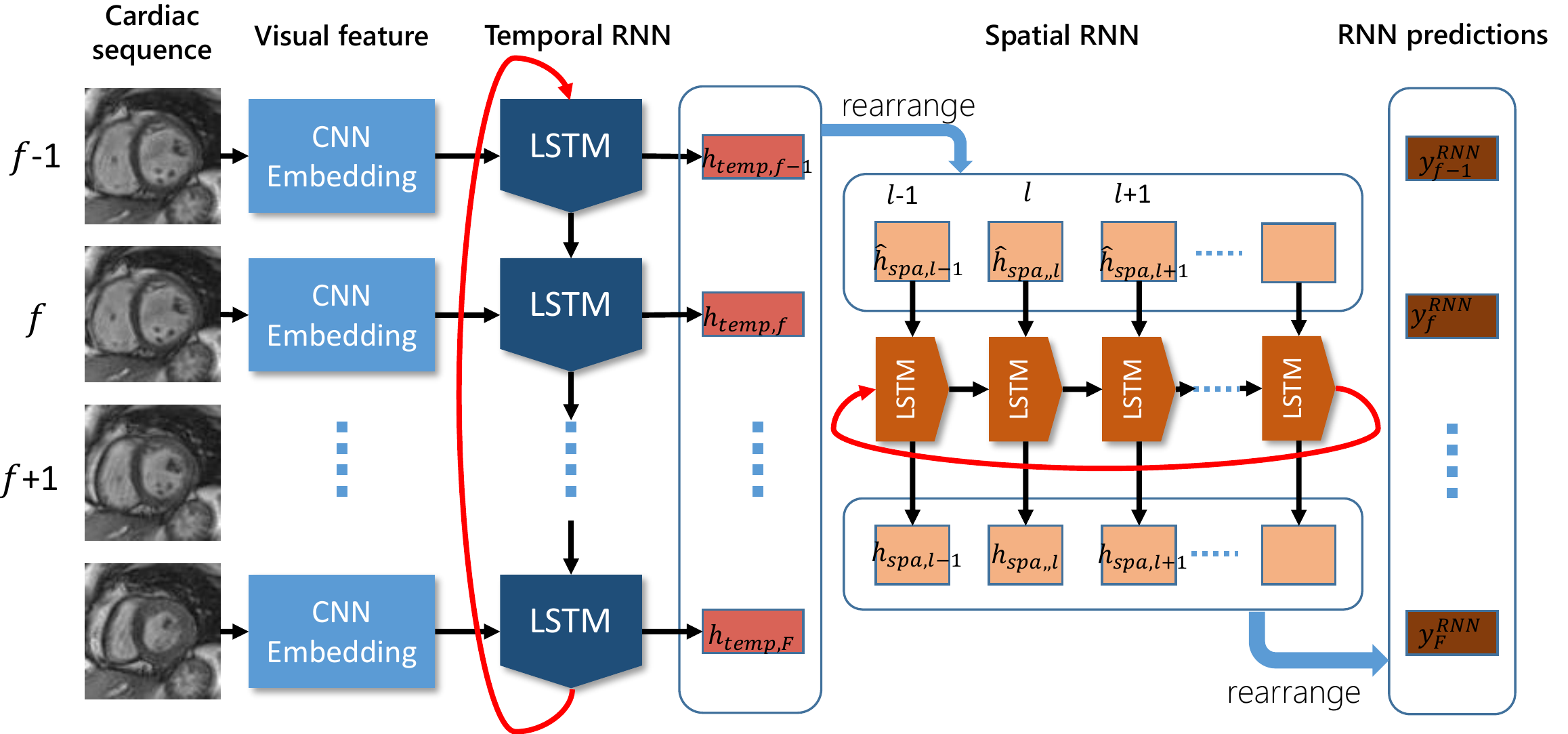}
	\caption{Diagram of the RNN path. The temporal dependencies among neighboring frames are modeled by the temporal-RNN with CNN embedding as input. The spatial dependencies among neighboring RWT are modeled by the spatial-RNN with rearranged output of temporal-RNN as input.}
	\label{fig_rnn_branch}
\end{figure}
The diagram of the RNN path is shown in Fig.~\ref{fig_rnn_branch}. Based on the CNN embedding obtained with the CNN path, temporal and spatial RNN are employed to effectively model the dependencies existing among the RWT of all frames. In this section, we first introduce the memory unit LSTM that we use in the RNN path, and then describe the temporal and spatial RNN.

\begin{figure}[t]
	\centering
	\includegraphics[width=7cm]{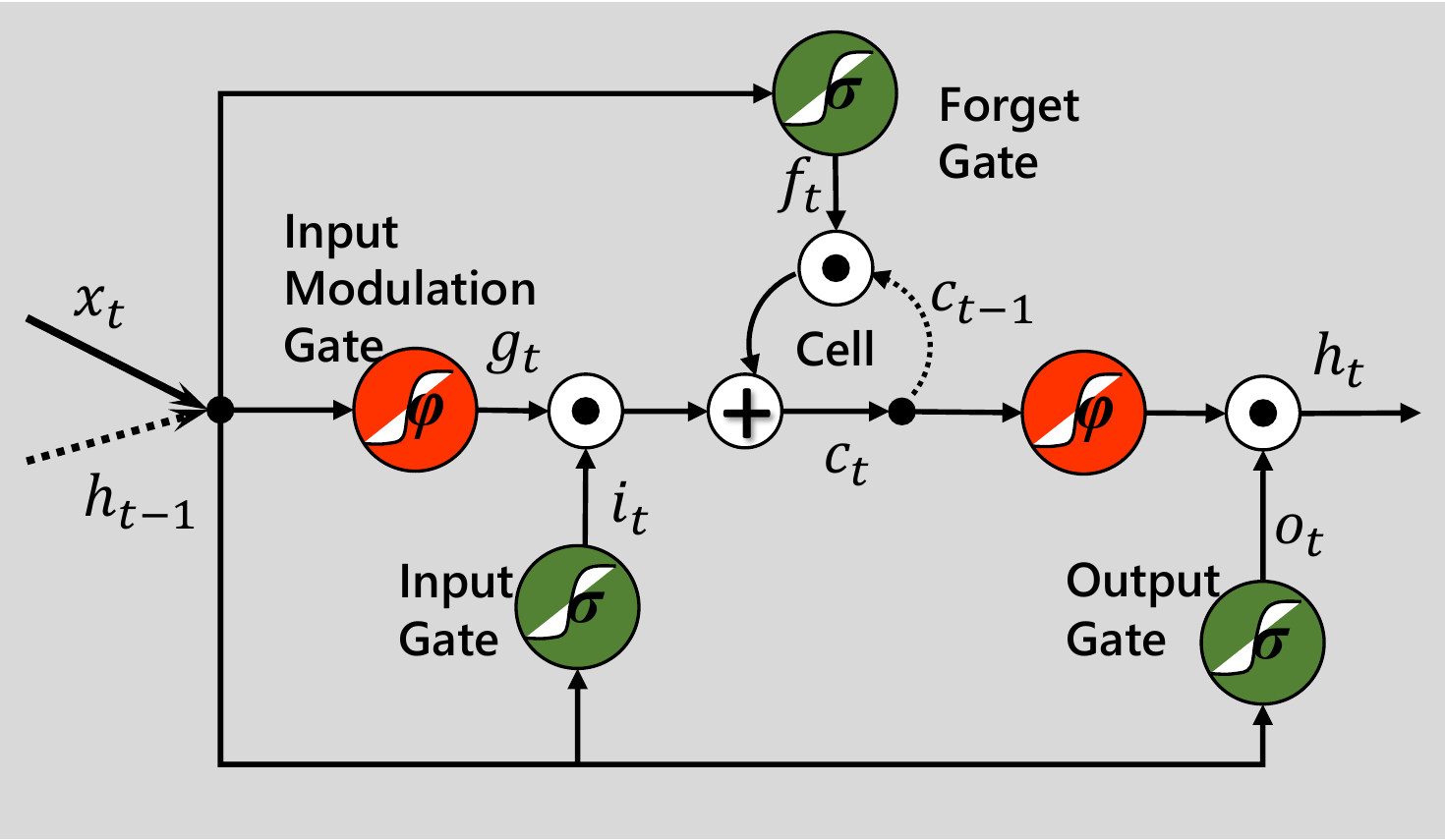}
	\caption{A diagram of LSTM unit in RNN, which is capable of adaptively modeling dynamic deformations in cardiac sequences due to the input gate, forget gate, and output gate.}
	\label{fig_lstm}
\end{figure}
\subsubsection{LSTM} In order to learn the long-term dynamics in sequential data and avoid the gradient vanishing/exploding problem in traditional RNN, LSTM unit~\cite{graves2012supervised}, as shown in Fig.~\ref{fig_lstm}, was introduced into RNN. The input gate, output gate, forget gate and the memory cell allow the network to learn when to forget previous hidden states and when to update current hidden states given current input. This strategy enables LSTM to adaptively memorize and access information long term ago. The LSTM computations for time step $t$ given the current input $x_t$, the previous hidden states $h_{t-1}$ and memory states $c_{t-1}$, are as follows~\cite{graves2012supervised}:
\begin{equation}
\begin{split}
i_t&=\sigma(W_{xi}x_t+W_{hi}h_{t-1}+b_i)\\
f_t&=\sigma(W_{xf}x_t+W_{hf}h_{t-1}+b_f)\\
o_t&=\sigma(W_{xo}x_t+W_{ho}h_{t-1}+b_o)\\
g_t&=\varphi(W_{xc}x_t+W_{hc}h_{t-1}+b_c)\\
c_t&=f_t\odot c_{t-1}+i_t\odot g_t\\
h_t&=o_t\odot\varphi(c_t)
\end{split}
\end{equation}
where $\sigma(\cdot)$ and $\varphi(\cdot)$ are element-wise sigmoid and hyperbolic tangent non-linearity functions, $\odot$ are element-wise product. The first three equations map the current input and previous hidden states to the input gate $i_t$, the forget gate $f_t$ and the output gate $o_t$, to adaptively control the information flow. $W$s are the weight matrices to be learned and $b$s are the corresponding bias terms.

\subsubsection{Temporal RNN and Spatial RNN}
With CNN embedding $e^s_f$, we first deploy a temporal RNN with the frame index in a cardiac sequences as time step to predict the values of RWT $h^s_{temp,f}\in \mathcal{R}^6$ for each frame $f$ taking account of the dependencies between neighboring frames (See \textit{Temporal RNN} in Fig.~\ref{fig_rnn_branch}). 
\begin{equation}
h^s_{temp,f} = LSTM(e^s_f, h^s_{temp,f-1}), f=1\dots F
\label{eq_temp_lstm}
\end{equation}
Based on the prediction of temporal RNN, we again deploy a spatial RNN with spatial location as time step to predict RWT  $h^s_{spa,l}\in \mathcal{R}^F$ of one specific location $l$ for all the frames in the sequences (See \textit{Spatial RNN} in Fig.~\ref{fig_rnn_branch}). We rearrange column vectors $[h^s_{temp,1}, h^s_{temp,2}, \dots, h^s_{temp,F}]$ to row vectors $[\hat{h}^{s,T}_{spa,1}; \hat{h}^{s,T}_{spa,2},\dots, \hat{h}^{s,T}_{spa,6}]$. Then we have
\begin{equation}
h^s_{spa,l} = LSTM(\hat{h}^s_{spa,l}, h^s_{spa,l-1}), l=1\dots 6
\label{eq_spa_lstm}
\end{equation}
By rearranging these row vectors $[h^{s,T}_{spa,1};h^{s,T}_{spa,2};\dots,h^{s,T}_{spa,6}]$ back to column vectors $[y^{s,RNN}_{1}, y^{s,RNN}_{2}, \dots, y^{s,RNN}_{F}]$, we get the RNN predictions for frame-wise RWT.
 
\subsubsection{Circle-RNN}
As can be seen from Eqs.~\ref{eq_temp_lstm} and~\ref{eq_spa_lstm}, three limitations exist for RNN: 1) for the first time step ($f=1$ or $l=1$), there is no value for previous hidden units, which influences the prediction of the first frame; 2) only past information can be used to determine the output of current time step, while future information is also equally important; 3) for the first few time steps, long-term dependency model can not be well built from the limited past information, leading to unfair treatment of different frames.

\begin{figure}[t]
	\centering
	\includegraphics[width=12cm,trim={1.8cm 0.2cm 1.5cm 0cm},clip]{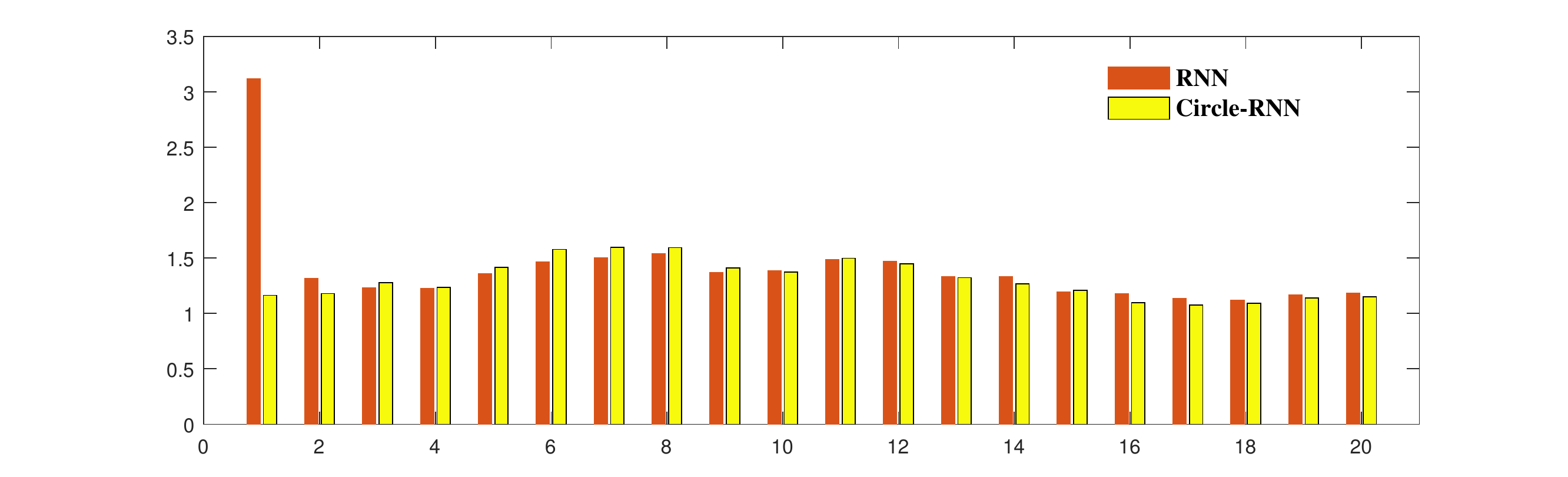}
	\caption{Average estimation error (mm) of different frames by RNN and Circle-RNN for cardiac RWT. Only temporal RNN is employed here for comparison. The estimation error for the first frame is clearly reduced by Circle-RNN.}
	\label{fig_bar_cirlstm}
\end{figure}

We provide an elegant solution to overcome these limitations for cardiac MR sequence: Circle-RNN, which connects the output of the last frame to the hidden input of the first frame, as the red arrows show in Fig.~\ref{fig_rnn_branch}. Within Circle-RNN, Eqs.~\ref{eq_temp_lstm} and~\ref{eq_spa_lstm} become: 
\begin{equation}
h^s_{temp,f} = LSTM(e^s_f, h^s_{temp,mod(f-1-1,F)+1}), f=1\dots F
\label{eq_temp_cir_lstm}
\end{equation}
\begin{equation}
h^s_{spa,l} = LSTM(\hat{h}^s_{spa,l}, h^s_{spa,mod(l-1-1,6)+1}), l=1\dots 6
\label{eq_spa_cir_lstm}
\end{equation}
where $mod(a,b)$ computes the modulus. Circle-RNN can be easily optimized with the BPTT algorithm~\cite{graves2012supervised}. To avoid infinite information loop within this Circle-RNN, we introduce a parameter \textit{depth} to control how many rounds the information flow in our Circle-RNN. Fig.~\ref{fig_bar_cirlstm} shows the error reduction of Circle-RNN over RNN when predicting cardiac RWT from the CNN embedding with only temporal RNN employed.

\section{Experiments}

\subsection{Dataset and Implementations}\label{sec_config}

A dataset of short-axis cine MR images paired with manually obtained ground truth values of RWT is constructed to evaluate the performance of our method, which includes 2900 images from 145 subjects. These subjects are collected from 3 hospitals affiliated with two health care centers and 2 vendors (GE and Siemens). Each subject contains 20 frames throughout a cardiac cycle. In each frame, the middle slice is selected following the standard AHA prescriptions~\cite{cerqueira2002standardized} for validation of the proposed cardiac RWT estimation method. The ground truth values of RWT are manually obtained for each image.

In our experiments, two landmarks, i.e, junctions of the right ventricular wall with the left ventricular, are manually marked for each cardiac image to provide reference for ROI cropping and the LV myocardial segments division. The cropped images are resized to dimension of $80\times80$. All values of RWT are normalized to the range of [0,1] according to the image dimension (80). Five-fold cross validation is employed for performance evaluation and comparison. Mean absolute error (MAE) between estimation and the ground truth is computed to evaluate the performance. The network is implemented by Caffe~\cite{jia2014caffe} with SGD solver. Learning rate and weight decay are set to (0.05, 0.0005). \textquoteleft step\textquoteright ~learning policy is employed with gamma and step size being (0.5, 2500) and momentum 0.9. The depth of Circle-RNN is set as the number of time steps, i.e, 20 for the temporal RNN and 6 for the spatial RNN. Data augmentation is conducted by randomly cropping images of size $75\times 75$ from the original image.

\subsection{Performance comparison: ResRNN vs. CNN and RNN}
\begin{table*}[t]
	\caption{Estimation error comparison of CNN, RNN and ResRNN for cardiac RWT estimation in terms of MAE (Mean$\pm$ standard deviation)(mm). WT-IS means wall thickness for the inferoseptal segment of myocardium, similar for the rest. Best results are highlighted in bold.}
	\label{table_analysis}
	\centering
	\begin{tabular}{l|cccc >{\bfseries}c}
		\hline
		Method & CNN &  RNN (plain) &RNN (circle)& ResRNN (plain)& ResRNN (circle)\\
		\hline
		WT-IS&1.28$\pm$1.10&1.30$\pm$1.11&1.31$\pm$1.04&1.26$\pm$1.15&1.22$\pm$1.02\\
		WT-I&1.56$\pm$1.27&1.60$\pm$1.28&1.54$\pm$1.26&1.55$\pm$1.24&1.47$\pm$1.22\\
		WT-IL&1.81$\pm$1.51&1.99$\pm$1.66&1.86$\pm$1.60&1.80$\pm$1.48&1.78$\pm$1.53\\
		WT-AL&1.64$\pm$1.30&1.70$\pm$1.42&1.66$\pm$1.35&\textbf{1.60$\pm$1.29}&1.60$\pm$1.31\\
		WT-A&1.35$\pm$1.09&1.36$\pm$1.18&1.30$\pm$1.09&1.40$\pm$1.20&1.31$\pm$1.08\\
		WT-AS&1.26$\pm$0.80&1.39$\pm$1.06&1.29$\pm$1.01&1.34$\pm$1.10&1.25$\pm$0.97\\
		\hline
		Average&1.49$\pm$0.76&1.56$\pm$0.83&1.49$\pm$0.80&1.49$\pm$0.81&1.44$\pm$0.74\\
		\hline
	\end{tabular}
\end{table*}

The advantages of the proposed ResRNN are firstly demonstrated by comparing performance of three different network architectures: 1) CNN, i.e. the CNN path as shown in Fig.~\ref{fig_cnn_branch}; 2) RNN, i.e. the RNN path as shown in Fig.~\ref{fig_rnn_branch}), and 3) the proposed ResRNN. For RNN and ResRNN, both the original RNN (the plain RNN) and Circle-RNN are employed for comparison. 

From the average estimation error shown in Table~\ref{table_analysis}, we can observe the followings. 1) The two-path ResRNN outperforms CNN and RNN, with either plain or circle RNN deployed, which can be ascribed to the complementarity of the preliminary estimation from each frame itself and the residual estimation that modeling the dependencies of cardiac sequence. 2) When Circle-RNN, rather than the plain RNN, is deployed, lower estimation error can be obtained by RNN or ResRNN, due to the fact that Circle-RNN is capable of memorizing useful cardiac dynamic information for the first time step. In the following experiments, Circle-RNN is deployed in ResRNN.

\subsection{Performance Comparison: ResRNN vs. state-of-the-art}

\begin{figure}[t]
	\centering
	\includegraphics[width=12cm,trim={1.8cm 0.2cm 1.5cm 0cm},clip]{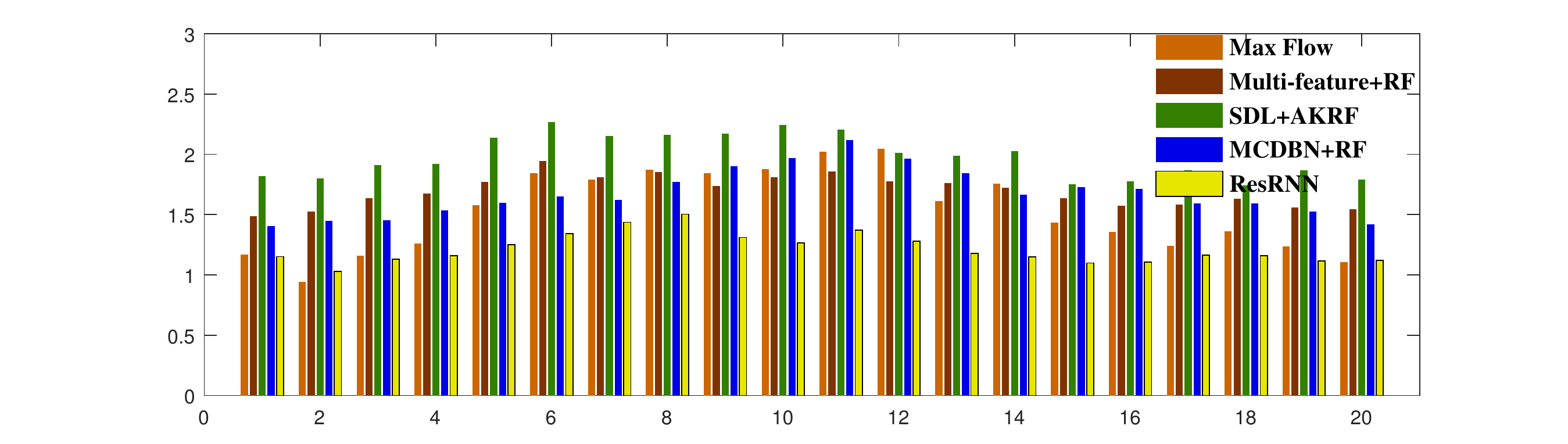}
	\caption{Average frame-wise estimation error (mm) of WT-IS over all subjects for ResRNN and all other competitors. The proposed ResRNN delivers clearly lower estimation error than other methods.}
	\label{fig_bar_error}
\end{figure}

To demonstrate the advantages of our proposed method over segmentation based~\cite{ayed2012max} and two-phase direct methods~\cite{zhen2014direct,zhen2015multi,zhen2015direct}, we apply these methods to our database for cardiac RWT estimation. For the direct methods, the same five-fold cross-validation protocol is used. As can be observed in Table~\ref{table_compare} and Fig.~\ref{fig_bar_error}, the proposed ResRNN demonstrates great advantages over existing segmentation-based and two-phase direct methods for cardiac RWT estimation. 

From Table~\ref{table_compare}, we can see that the proposed ResRNN estimates cardiac RWT with high accuracy (average MAE of 1.44mm) and outperforms all competitors. Specifically, it outperforms the Max Flow method with 55.14\% MAE reduction. Note that Max Flow obtained high dice metric of 0.9182 for LV cavity segmentation when applied to our database. In fact, the dependency on manual segmentation of the first frame makes the estimation error of Max Flow increase as the estimated frame is far from the first frame within the cardiac cycle (see the frame-wise estimation error of Max Flow in Fig.~\ref{fig_bar_error}). ResRNN outperforms the best of the direct methods (MCDBN+RF~\cite{zhen2015multi}) with a clear error reduction: 12.73\%. Within the two-phase framework, the hand-crafted multifeatures, the features obtained by supervised learning, and MCDBN features employed in existing methods all fail to beat ResRNN, which evidences the benefits of the network architecture in ResRNN over the two-phase direct methods. Fig.~\ref{fig_bar_error} also shows that ResRNN estimates each frame with consistently lower MAE.

\begin{table*}[t]
	\caption{Estimation error comparison of the proposed ResRNN with segmentation based method and existing direct methods for RWT estimation in terms of MAE (Mean$\pm$ standard deviation)(mm). Best results are highlighted in bold.}
	\label{table_compare}
	\centering
	\begin{tabular}{l|ccccc}
		\hline
		\multirow{2}{*}{Method} & Max Flow & Multi-features+RF& SDL+AKRF & MCDBN+RF& \multirow{2}{*}{ResRNN}\\
		&\cite{ayed2012max}&~\cite{zhen2014direct}&~\cite{zhen2015direct}&~\cite{zhen2015multi}&\\
		\hline
		\hline
		WT-IS&1.53$\pm$1.73&1.70$\pm$1.47&1.98$\pm$1.58&1.78$\pm$1.40&\textbf{1.22$\pm$1.02}\\
		WT-I&3.23$\pm$2.83&1.71$\pm$1.34&1.67$\pm$1.40&1.68$\pm$1.41&\textbf{1.47$\pm$1.22}\\
		WT-IL&4.15$\pm$3.17&1.97$\pm$1.54&1.88$\pm$1.63&1.92$\pm$1.45&\textbf{1.78$\pm$1.53}\\
		WT-AL&5.08$\pm$3.95&1.82$\pm$1.41&1.87$\pm$1.55&1.66$\pm$1.20&\textbf{1.60$\pm$1.31}\\
		WT-A&3.47$\pm$3.25&1.55$\pm$1.33&1.65$\pm$1.45&\textbf{1.20$\pm$1.01}&1.31$\pm$1.08\\
		WT-AS&1.76$\pm$1.80&1.68$\pm$1.43&2.04$\pm$1.59&1.63$\pm$1.23&\textbf{1.25$\pm$0.97}\\
		\hline
		Average&3.21$\pm$1.98&1.73$\pm$0.97&1.85$\pm$1.03&1.65$\pm$0.77&\textbf{1.44$\pm$0.74}\\	
		\hline
	\end{tabular}
\end{table*}

Besides, we can draw that all the deep neural network based methods in Table~\ref{table_analysis} outperform existing two-phase direct methods in Table~\ref{table_compare}. This further confirms the argument that independent feature extraction and regression cannot make the two phases maximumly compatible with each other. The end-to-end learning procedure of neural network integrates both phases together and leads to clearly improved performance.

\section{Conclusions}
In this paper, we propose an effective network architecture ResRNN for the task of cardiac RWT estimation, which has never been explored before. In ResRNN, a CNN path is employed to estimate from each cardiac image independently the preliminary results of RWT, and an RNN path is employed to compensate the residual of CNN estimations with temporal and spatial dependencies being accounted by Circle-RNN. Validation on a data set of cardiac MR sequences from 145 subjects demonstrates that the proposed ResRNN is capable of estimating cardiac RWT values with performance better than state-of-the-art methods, and is of great potential in clinical cardiac function assessment.

\bibliographystyle{splncs03}
\bibliography{cirrnn_wt_short}

\end{document}